\def\year{2017}\relax
\documentclass[letterpaper]{article} 
\usepackage{aaai17}  
\usepackage{times}  
\usepackage{helvet}  
\usepackage{courier}  
\usepackage{relsize}
\usepackage[hyphenbreaks]{breakurl}
\usepackage[hyphens]{url}
\usepackage{graphicx}  
\usepackage[nounderscore]{syntax}
\usepackage{newfloat}

\usepackage[export]{adjustbox}
\usepackage[nounderscore]{syntax}
\usepackage{newfloat}
\DeclareFloatingEnvironment[
  fileext   = logr,
  listname  = {List of Grammars},
  name      = Grammar,
  placement = htp
]{Grammar}

\begin{document}
%
\title{``Press Space To Fire'': Automatic Video Game Tutorial Generation}
\author{
Michael Cerny Green, Ahmed Khalifa, Gabriella A. B. Barros, \and Julian Togelius\\
Tandon School of Engineering,
New York University,
New York, USA\\
mcg520@nyu.edu, aak538@nyu.edu, gabbbarros@gmail.com, julian@togelius.com
}
\maketitle
\begin{abstract}
We propose the problem of tutorial generation for games, i.e. to generate tutorials which can teach players to play games, as an AI problem. This problem can be approached in several ways, including generating natural language descriptions of game rules, generating instructive game levels, and generating demonstrations of how to play a game using agents that play in a human-like manner. We further argue that the General Video Game AI framework provides a useful testbed for addressing this problem.
\end{abstract}

\section{Introduction}

Artificial intelligence techniques can be used in and with games in many different ways, to solve problems and create experiences, as well as to advance AI. Very coarsely, AI can be applied to generate content \cite{shaker2016procedural}, play games, and model players~\cite{yannakakis2017artificial}, though there are many examples of usage of AI which do not fit these categories cleanly. Looking at AI-based game design patterns, one can see AI occasionally being used in somewhat more obscure roles such as spectacle, trainee or co-creator~\cite{treanor2015AI}.

With this paper, we seek to introduce yet another interesting problem, and role, for AI in games. The problem is that of generating tutorials (or instructions) and the role is that of teacher. We can loosely define it as: \emph{given a game, generate a way to teach players how to play it}.

Most video games feature some kind of tutorial or instructions to assist players in getting started, and creating such tutorials is a complex task that requires skill and time. In other words, it is a great candidate for total or partial-automation. 
Furthermore, if we can find good methods for generating tutorials, these methods can be used to help teach people to perform a large variety of other tasks.

This paper surveys the (scant) literature on game tutorials and makes a few basic distinctions between types of tutorials. We then discuss some possible approaches to tutorial generation---it turns out this problem has much in common with, and builds on advances in, human-like game playing as well as procedural content generation~\cite{shaker2016procedural}. Finally, we discuss what it would take to generate tutorials within the General Video Game AI (GVG-AI) framework.

\section{Background}
Tutorials are the first interactions players encounter in a game. They help players understand game rules and, ultimately, learn how to play with them. In the game industry, developers experimented with different tutorial formats ~\cite{therrien2011get}. In the arcade era, when most games were meant to be picked up and played quickly, they either had very simple mechanics, or they contained mechanics that players could relate to: ``Press right to move'', ``Press up to jump'', and so on. As a result, these games usually lacked a formal tutorial. As their complexity increased and home consoles started to explode in popularity, formal tutorials became more common. 

Some game developers tried using an active learning approach which was optimized for players that learn through experimentation and exploring carefully designed levels. Games like \emph{Megaman X} (Capcom, 1993) follow this approach. Other developers relied on old-school techniques, teaching the player everything before they could play the game, such as in \emph{Heart of Iron 3} (Paradox Interactive, 2009). While one cannot argue that one technique is always superior to another, different techniques suit different audiences and/or games~\cite{andersen2012impact,williams2009pedagogy,ray2010learning}.

Tutorials have evolved significantly over time, from the simple directive of Pong (``Avoid missing the ball for highscore'') to the exquisitely detailed in-game database of Civilization \cite{therrien2011get}. Suddaby describes multiple types of tutorials \cite{suddaby2012many}, from none at all to \textit{thematically relevant contextual lessons}, where the tutorial is ingrained within the game environment.

Tutorial types are related to the different learning capabilities of the users who play them. Sheri Graner Ray~(\citeyear{ray2010learning}) discusses different \textit{knowledge acquisition styles} in addition to traditional learning styles: Explorative Acquisition and Modeling Acquisition. The first style incorporates a child-like curiosity and ``learning by doing'', whereas the second is about knowing how to do something before doing it. We can define at least two distinct tutorial styles from this, one being exploratory during gameplay and the other being more instructional before the game even begins.

Williams suggests that active learning tutorials, which stress player engagement and participation with the skills being learned, may be ineffective when the player never has an isolated place to practice a particularly complex skill~\cite{williams2009pedagogy}. In fact, Williams argues that some active learning tutorials actually ruin the entire game experience for the player because of this reason.
According to Andersen et al., the effectiveness of tutorials on gameplay depends on how complex a game is to begin with~\cite{andersen2012impact}, and sometimes are not useful at all. Game mechanics that are simple enough to be discovered using experimental methods may not require a tutorial to explain them. From these two sources, we find our first two boundaries for tutorial generation: there exists mechanics that are too simple to be taught in a tutorial, and there are mechanics complex enough that they may need to be practiced in a well-designed environment to hone.

In general, a game developer would want to use the most suitable tutorial style for their game. For that purpose, they must understand different dimensions/factors that affect the tutorial design process and outcome. Andersen et al.~\cite{andersen2012impact} measured how game complexity affects the perceived outcome of tutorials. In their study, they defined 4 dimensions of tutorial classification: 
\begin{itemize}
\item \textbf{Tutorial Presence:} whether the game has a tutorial or not.
\item \textbf{Context Sensitivity:} whether the tutorial is a part of story and game or separate and independent from them.
\item \textbf{Freedom:} whether the player is free to experiment and explore or is forced to follow a set of commands.
\item \textbf{Availability of Help:} whether the player can request for help or not.
\end{itemize}

\begin{figure}[h]
\centering
\includegraphics[width=0.9\columnwidth]{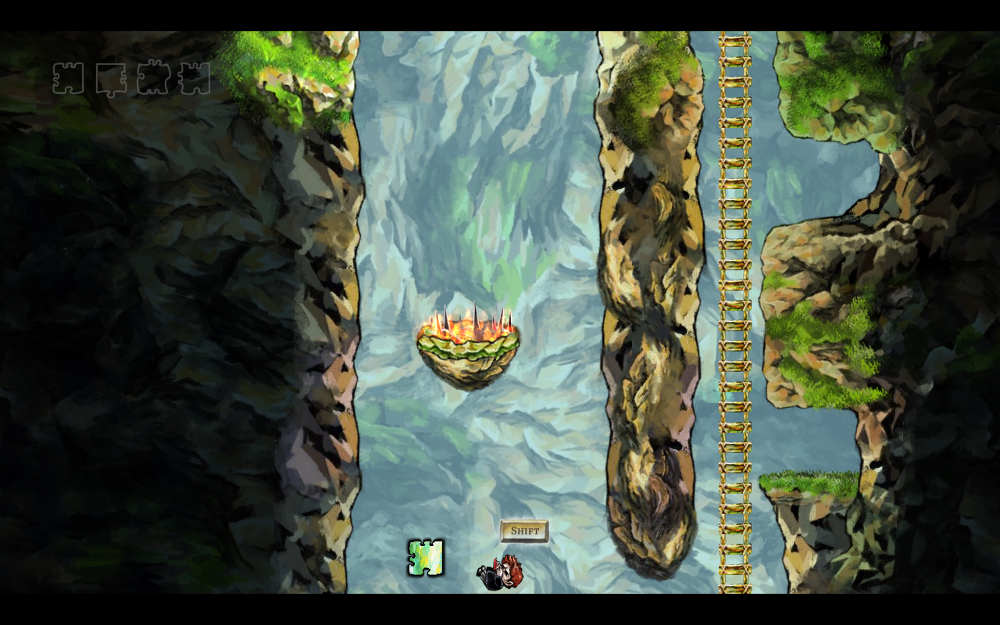}
\caption{Braid teaching time rewinding mechanic when the player dies.}
\label{fig:braid}
\end{figure}

The classification proposed by Andersen et al. is binary.
However, it is useful to see tutorials situated on a continuum between these extremes, as this allows us to gain a more nuanced understanding of game tutorials. For example: Figure ~\ref{fig:braid} shows the tutorial in \emph{Braid} (Number None, Inc, 2008) for a time rewinding mechanic. The tutorial only appears based on a certain event, i.e. the player's death. Players will not know about the mechanic until their first death. Instead of having the tutorial available at anytime or showing how to use the mechanic at the beginning, the developer reveals it when it is first necessary. 

Sampling this space and comparing it with current game tutorials, we can find patterns repeated in multiple games. We can highlight the following tutorial types, which are not the only tutorials present in the space, but appear to be the most common ones:
\begin{itemize}
\item \textbf{Teaching using instructions:} These tutorials explain how to play the game by providing the player with a group of instructions to follow, similar to what is seen in boardgames. For example: Strategy games, such as \emph{Starcraft} (Blizzard, 1998), teach the player by taking them step by step towards understanding different aspects of the game.
\item \textbf{Teaching using examples:} These tutorials explain how to play by showing the player an example of what will happen if they do a specific action. For example: Megaman X uses a Non Playable Character (NPC) to teach the player about the charging skill~\cite{egoraptor2011megaman}.
\item \textbf{Teaching using a carefully designed experience:} These tutorials explain how to play the game by giving the player freedom to explore and experiment. For example: in \emph{Super Mario Bros} (Nintendo, 1985), the world 1-1 is designed to introduce players to different game elements, such as goombas and mushrooms, in a way that the player can not miss~\cite{credits2014supermario}. One way of seeing that is that early obstacles are instances of \emph{patterns}, which reoccur later in the game in more complex instantiations or combinations~\cite{dahlskog2012patterns}.
\end{itemize}

\begin{figure}[h]
\centering
\includegraphics[width=0.9\columnwidth]{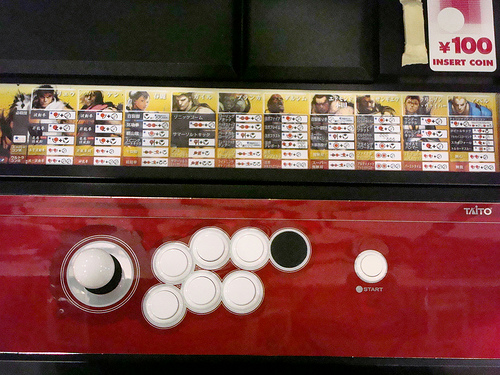}
\caption{Street Fighters arcade cabinet. The cabinet shows different combos that can be done.}
\label{fig:streetfighter}
\end{figure}

A game can have more than a single tutorial type from the previous list. Arcade games used demos and instructions to both catch the attention of the player and help them learn it. The demos help to attract more players, while simultaneously teaching them how to play. On the other hand, showing an information screen before the game start, such as in \emph{Pacman} (BANDAI NAMCO, 1980), or displaying instructions on the arcade cabin, frequently seen in fighting games, helps the player understand the game and become invested in it. Figure \ref{fig:streetfighter} shows \emph{Street Fighters} arcade cabinet where different characters combos and moves are written on it. Megaman X uses a carefully designed level to teach the player what to do, but still gives an example of the powershot attack if the player missed it.

Previous work has been done related to tutorial generation, especially in the area of aiding beginners, such as in Blackjack heuristics \cite{de2016generating}, by evolving the heuristics to be effective and concise. Similarly, TutorialPlan~\cite{li2013tutorialplan} generates text and image instructions for users to learn AutoCAD.
Work has also been done in automatic tutorial generation for API coding libraries~\cite{単体テストを用いたチュートリアルの自動生成手法2014automatic}, which claimed that the resulting generated tutorials helped users learn libraries more effectively than current auto-generated tutorials. Alexander et al. postulated that open world mechanics could be transformed into quests \cite{alexander2017deriving}. By formalizing the game logic of Minecraft into rules, they were able to create action graphs representing the player experience, and create quests and achievements based off those actions.



Game-O-Matic~\cite{treanor2012game} is a system which generates arcade style games and instructions for them by using a story-based concept-map inputted by a user. After the game is created, Game-0-Matic generates a tutorial page, explaining who the player will control, how to control them, and winning/losing conditions, by using the concept-map and relationships between objects within it.

\section{How could tutorials be generated?}

There already exist numerous artificial agents for a variety of video games. We theorize that if an agent can beat a video game, it could also build a tutorial using parts of the methodology that it used to win. An agent could do so by constructing a game-mechanic graph, similar to the mission-graph as described by Dormans~\cite{dormans2010adventures}.

With the knowledge of what it takes to beat a level (or to accomplish a specific goal), the agent could construct the graph so that leaf nodes contain terminal states, such as finishing a side quest or beating the level. Nodes in between the initial state and a terminal one would contain mechanics that lead to a terminal state such as player death or winning the game. 

\begin{figure}[h]
\includegraphics[width=\columnwidth]{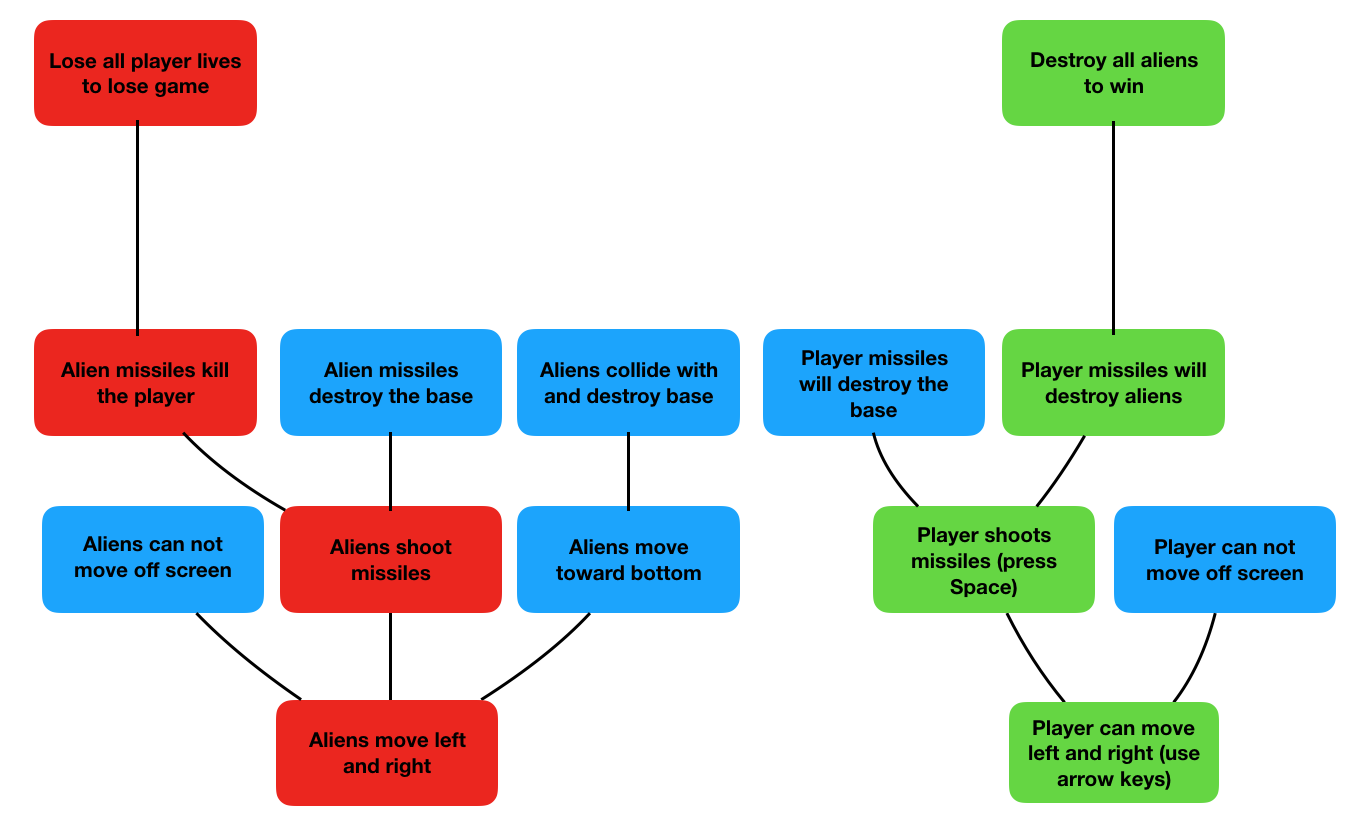}
\caption{A Space Invaders tutorial graph. Green nodes are on a critical victory path. Red nodes are on a critical loss path}
\label{fig:space-invaders}
\end{figure}

Figure \ref{fig:space-invaders} displays a tutorial graph created for \emph{Space Invaders} (Taito, 1978). The green nodes are actions that equate to a critical victory path. Doing these actions will eventually result in victory. Red nodes are a critical loss, and doing these actions could result in losing the game. All nodes could be transformed into steps in a tutorial. The steps will explain how to achieve victory, avoid loss, or other subgoals such as how to score points. It is important to mention that nodes not on a critical path of either type can optionally be present in a tutorial and are not necessary.

The graph is flexible enough to be incorporated into different types of tutorial methods, for example those described in the previous section using \textit{instruction}, \textit{examples}, and \textit{a carefully designed experience} for Space Invaders:
\begin{itemize}
\item \textbf{Tutorial using instructions:} Using the text in the node as a start, a grammar could piece together sentences explaining each critical node on the winning and losing paths. The user would see text with phrases like ``Press the left and right arrow keys to move left and right.''
\item \textbf{Tutorial using examples:} Since the graph was built by an agent, we can assume the agent learned these mechanics and can replicate them. The tutorial generator could build an example by creating a level stage that would isolate the action or behavior contained within each node. The user would see a human-like artificial agent, such as one created by Khalifa et al. \cite{khalifa2016modifying}, using the isolated movement mechanic, moving left and right on the screen. Using a human-like AI should help ensure that it would take similar actions to that of a human being, so that the observing player might learn more effectively.
\item \textbf{Tutorial using a carefully designed experience:}  This combines the \textit{tutorial using examples} experience with levels that revolve around the mechanics to be learned. Rather than an artificial agent playing, the player would be in control. A level generator would be generating levels based around the critical path nodes. One such designed level would introduce player movement and shooting, and eventually move up to shooting aliens, finally teaching that killing all aliens will win the level, whilst still allowing the player to move around freely and unrestricted.
\end{itemize}

A tutorial generation system for teaching using instructions would require some sort of text generation engine, which would create text to be shown to the player to teach them how to play. The most obvious choice for this would be a grammar-based system, as it would allow the greatest amount of flexibility, especially over a simple text-replacement system. Grammars have been shown to be highly effective generators of content in the past~\cite{rumelhart1975notes,pemberton1989modular,dormans2011level,dormans2011generating,dart2011speedrock,togelius2016grammars,callaway2002narrative}.
One such example of a grammar-based text generative tool is Tracery~\cite{compton2014tracery}, which was made to generate stories with a pinch of the nonsensical. Stories generated by Tracery are not bound to be causal or even make much sense. Past research~\cite{lang1999declarative} has shown that a system that generates sophisticated story-lines requires massive amounts of meta-data, which is often unwieldy, expensive in overhead, and arguably inflexible. Luckily, a tutorial generation system would not be required to tell a story, but rather teach the player how to play a game. Therefore, we believe that a grammar-based generator will suffice, even if it does not write causally. 

Another way to generate text is using an artificial agent would explain its actions as it plays the game. Schrodt et al created an artificial agent to play Super Mario, which literally "thought-out-loud" as it played the game~\cite{schrodt2017event}.  This technique of voicing intent or decision during play is known as \emph{framing} \cite{charnley2012notion}. The human-like AI mentioned before could be modified to explain its decision-making in real-time in order to teach a human-player how to play. 

\section{GVG-AI Tutorial Generation}

The GVG-AI Framework provides a testbed for researchers to solve the problem of general video game-playing artificial intelligence, a competition where competitors can design AI agents that play a variety of unseen games efficiently.
Multiple competition tracks are available to compete in, including Agent \cite{perez20162014}, Level Generation \cite{khalifa2016general}, Multiplayer Planning \cite{gaina2016general}, Learning, and Rule Generation.

In this paper, we propose a method of tutorial generation and provide a possible beginning for a tutorial generation track for GVG-AI. 

Because of the wide-ranging nature of game types in the GVG-AI framework, generated tutorials would have to be applicable to a variety of game styles and mechanics. As all games in GVG-AI are written in VGDL, tutorial generation can be done by simply reading the various interactions and terminations in the game's VGDL file and translating it into an easy-to-read, concise format. Before the first level of the game begins, a text display demonstrating button usage, enemy types, the player, and collectibles would be shown, as well as pointing out the main goal of the game.
Tutorial generation can be divided into \textit{mechanic discovery},
\textit{graph creation}, and \textit{text generation}, all of which will be described in the following subsections.

\subsection{Mechanic Discovery}
To generate a tutorial, our engine must first learn all relevant game mechanics. Using the \textit{Sprite}, \textit{Interaction}, and \textit{Termination} sets contained within the game's VDGL file, the engine would record interactions between various sprites, movement controls, and terminal states. 
For example, Figure \ref{fig:Aliens-Ruleset} shows the interaction and termination sets for Space Invaders.  The \textit{avatar} is defined in the SpriteSet to be a FlakAvatar, which means it can shoot \textit{missiles}. In the InteractionSet, missiles that collide with EOS (End of Screen) are destroyed.

An alternative to reading the interaction rules is using artificial agents to play the game. An AI agent would explore the game space, and discover game mechanics on its own. 

\subsection{Build Graph}
The engine must have some way of understanding and relating game mechanics to one another. We propose a graph-based rule-set interpretation.
The engine would create a graph based off the discovered mechanics, where each node contains a mechanic of some kind. Nodes that are connected to each other are related in some way. Leaf nodes can contain terminal states for the game, such as win or loss. If the terminal state is a winning state, this would become a potential candidate for the \textit{critical path}, the shortest interaction chain necessary to win the level out of all interaction chains. This path would be the key goal the generator would recommend doing. Other paths that include losing states would correspond to things the generator would recommend \textit{not} to do. Paths that have score changes, either in a positive or negative way, would correlate to actions the generator would recommend to either do often or avoid doing as much as possible. Multiple critical paths might result from this, which means the generator would have to find the most efficient critical path. 

Alternatively, the engine could use a human-like AI \cite{khalifa2016modifying} armed with the knowledge of these discovered mechanics. This agent could play the game similarly to a human player to discover the critical path. Any game mechanics that it associates with loss would be placed on a critical loss path, and any game mechanics that it associates with winning would be placed on a critical win path.

\begin{figure} [t]
\includegraphics[width=\columnwidth, frame]{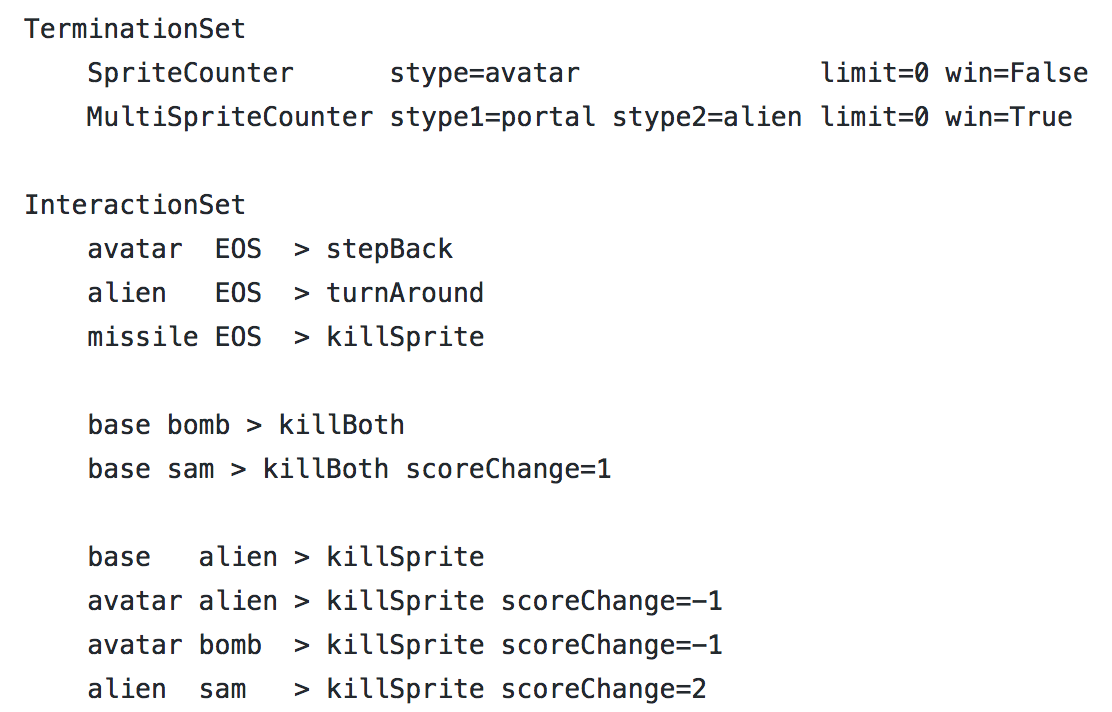}
\caption{Part of the rules for Aliens in VGDL}
\label{fig:Aliens-Ruleset}
\end{figure}

Here is an example of how the engine would build an interaction chain. In GVG-AI the arrow keys are assumed to be the movement controls. Referencing Figure \ref{fig:Aliens-Ruleset}, the avatar (the player) is defined to be a "FlakAvatar" which is restricted to only horizontal movement. Thus the first green node would be created with references to "move", "player", and "left-right" movement using "arrow keys". Now within the interaction set, the system would first look for associations to "avatar" to add to the chain. It would see the first interaction "avatar EOS $>$ stepBack", which means the avatar can not move outside the bounds of the screen. A node would be created with associations to "player", "not move" and "EOS" and linked to the previous node, as they are related via "avatar" and "movement".

\subsection{Generate Text with Grammar}
\begin{Grammar} [t]
\hrule
\begin{grammar}
<tutorial> ::= <win> <lose> <negative> <positive> <mechanics> <controls>

<win> ::= 'To win' <actionVerb> <helpingAdj> <sprite>

<lose> ::= 'To lose' <actionVerb> <helpingAdj> <sprite> | $\epsilon$

<negative> ::= 'Avoid' <actionVerb> <sprite> <negative> | $\epsilon$

<positive> ::= <actionVerb> <sprite> <positive> | $\epsilon$

<mechanics> ::= <mech> <mechanics> | <mech>

<mech> ::= <sprite> <helpingVerb> <actionVerb> <helpingAdjective> <sprite>

<controls> ::= <cont> <controls> | <cont>

<cont> ::= <controlVerb> <button> 'to' <actionVerb> <sprite>

<helpingAdj> ::= 'all' | 'every' | 'one' | 'some' | 'none' | $\epsilon$

<helpingVerb> ::= 'can' | 'can not' | 'will' | 'will not' | $\epsilon$

<actionVerb> ::= 'move' | 'shoot' | 'dodge' | 'kill' | 'destroy' | 'collide with' | 'beat' | 'lose' | 'win'

<controlVerb> ::= 'press' | 'hold' | 'release'

<button> ::= 'arrow keys' | 'left and right' | 'space bar'
\end{grammar}
\hrule
\caption{An example grammar for GVG-AI games.}\label{fig:grammar}
\end{Grammar}

A tutorial is not complete without some vehicle through which to educate the player about the game. Using the graph of game mechanics, the engine would generate a text-based tutorial to display instructions to the player using a grammar for building blocks. Grammar \ref{fig:grammar} displays an example grammar that could be used for a GVG-AI Tutorial Generator. Using the grammar, the system would designate which of the sentence types are applicable for a given node. For example. The bottom-right green node ("Player can move left and right (using arrow keys)") references a behavior about movement (an Action Verb), the player (a Sprite), and infers arrow key controls (Control Verb and Button). Thus the system would determine that a Control Sentence would be most applicable to this node, and form the sentence "Press arrow keys to move player". The second bottom-most red node ("Aliens shoot missiles") would be determined to be a Mechanic Sentence, as it references behaviors pertaining to Alien and Alien Missile (two Sprites) and "shoot" (an Action Verb). Thus the engine would build a sentence "Alien shoots alien missile".  Of important note, $<$sprites$>$ are not defined in the grammar, as they differ between games in GVG-AI. Space Invaders has sprites such as ``Alien'' and ``Missile'' whereas Solarfox contains ``Blib'' and ``PowerBlib''.

\section{Conclusion}

Ideas behind designing game tutorials have evolved over time. It is ironic that, while so much effort has been put into generating levels, textures and stories, little to no effort has been made into automatically generating what is the first interaction between players and the gameplay. This paper proposes the problem of automatically generating game tutorials through the lenses of an AI problem. It expands on the four dimensions defined by Andersen et al \cite{andersen2012impact} for classifying different types of tutorials, and highlights three mainstream tutorial types: Teaching using instructions, using examples and using a carefully designed experience. It is our belief that the GVG-AI framework can be a useful testbed for experimenting with tutorial generation. Finally, we propose a graph-based rule-set interpretation of an agent playthrough, presented to the player as a grammar-based text tutorial. As it is, that proposal is our first future step.

\bibliography{bibliography}
\bibliographystyle{aaai}
\end{document}